\documentclass[journal]{IEEEtran}
\usepackage{graphicx}
\usepackage{comment}
\usepackage{amsmath,amssymb} 
\usepackage{color}
\usepackage{enumitem}
\usepackage{booktabs}
\usepackage{tabu}
\usepackage{multirow}
\usepackage{setspace}
\usepackage[outdir=./figs/]{epstopdf}
\usepackage[linesnumbered,ruled]{algorithm2e}
\usepackage[pagebackref=true,breaklinks=true,letterpaper=true,colorlinks,bookmarks=false]{hyperref}

\begin{document}
\title{Jointly Modeling Motion and Appearance Cues for Robust RGB-T Tracking}

\author{Pengyu Zhang, \ \ \ Jie Zhao, \ \ \  Dong Wang, \ \ \  Huchuan Lu, \ \ \ Xiaoyun Yang
}

\markboth{IEEE Transactions on Image Processing}%
{Wang~\MakeLowercase{\textit{et al.}}: Online Object Tracking with Sparse Prototypes}

\maketitle

\begin{abstract}
In this study, we propose a novel RGB-T tracking framework by jointly modeling both appearance and motion cues. 
First, to obtain a robust appearance model, we develop a novel late fusion method to infer the fusion weight maps of both RGB and thermal (T) modalities. 
The fusion weights are determined by using offline-trained global and local multimodal fusion networks, and then adopted to linearly combine the response maps of RGB and T modalities. 
%
Second, when the appearance cue is unreliable, we comprehensively take motion cues, i.e., target and camera motions, into account to make the tracker robust.
%
We further propose a tracker switcher to switch the appearance and motion trackers flexibly. 
Numerous results on three recent RGB-T tracking datasets show that the proposed tracker performs significantly better than other state-of-the-art algorithms.
\end{abstract}


\section{Introduction}\label{sec1}
\begin{figure}[t]
	\centering
	\footnotesize
	\includegraphics[width=1.0\linewidth]{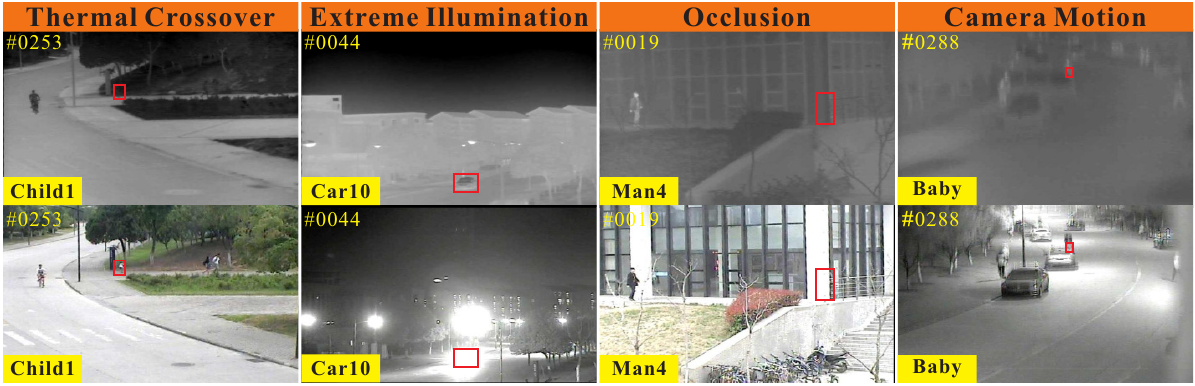}
	(a)~~~~~~~~~~~~~~~~~~~~(b)~~~~~~~~~~~~~~~~~~(c)~~~~~~~~~~~~~~~~~~~~(d)

	\caption{Tracking with RGB-T modalities may suffer from thermal crossover, extreme illumination, occlusion, 
		camera motion, and so on.}
	\label{fig:fig1}
\vspace{-5mm}
\end{figure}

RGB-T tracking aims to integrate complementary visible (RGB) and thermal (T) infrared information to boost the tracking performance and make the tracker work in day and night~\cite{Thermal-Survey}. 
First, the thermal infrared information is insensitive to illumination conditions and captures the target in extreme weather conditions, including night, fog, smog, to name a few. 
Second, the visible information is more discriminative in foreground-background separation under normal circumstances and is more effective in separating two moving targets when thermal crossover occurs. 
Although many works have been done in recent years~\cite{CFM,TFF,DAPNet,MANet,DAFNet}, 
effective fusion of RGB and T modalities and exploration of motion cues have big potential in designing a robust RGB-T tracker.
Multimodal fusion is crucial to develop a robust appearance model in the RGB-T tracking task. 
As shown in Figure~\ref{fig:fig1}, information from each single modality is not always reliable, 
due to thermal crossover and extreme illumination. 
Existing methods mainly focus on information aggregation from different modalities.
Li~\emph{et al.}\cite{MANet} propose MANet to integrate both modality-shared and modality-specific 
information in an end-to-end manner.
Zhang~\emph{et al.}\cite{mfDiMP} extend an RGB tracker to handle the RGB-T tracking task, 
and analyze different multimodal fusion types.
Li~\emph{et al.}\cite{CMR} focus on eliminating the modality discrepancy to exploit 
different properties between two modalities at the feature level. 
All aforementioned methods belong to early fusion, the strength of late fusion has not been explored 
in RGB-T tracking. 
In this work, we develop a novel late fusion method to generate the fusion weights of RGB and T 
modalities and use them to fuse the response maps for robust tracking. 
Motion information is also very important especially when the appearance cue is unreliable. 
Figure~\ref{fig:fig1} shows that the target appearance dramatically changes when occlusion and 
camera motion occur. 
First, the appearance information is meaningless when the target is fully occluded (Figure~\ref{fig:fig1} (c)).
Second, camera moving with motion blur and out of search region makes the appearance model 
less effective (Figure~\ref{fig:fig1} (d)).
In aforementioned situations, we resort to some motion models for target prediction 
and camera motion compensation. 
Motion models have been exploited for traditional RGB tracking to deal with either target 
motion~\cite{PF-Li08,GeometricPF-Kwon} or camera motion~\cite{UAV70}, but 
few works have been done in the RGB-T tracking task. 
In this work, we attempt to effectively model target and camera motions, acting as a vital supplement to the appearance model. 

Motivated by the aforementioned discussions, this work attempts to jointly model the appearance and motion 
information for robust RGB-T tracking. Our contributions are summarized as follows.

\begin{itemize}[itemsep = 3 pt,topsep = 3 pt, parsep = 2 pt, partopsep = 1pt ]
	\item We propose a novel RGB-T tracking framework to take both appearance and motion information 
	into account,  thereby resulting in a very robust performance.    
	\item We propose a novel late fusion method (MFNet) to obtain both global and local weights for 
	effective multimodal fusion, thereby resulting a robust response map. 
	\item We exploit both camera motion and target motion to mine the motion cues for RGB-T tracking and propose a new scheme to dynamically
	switch between appearance and motion cues.
	\item  Extensive experiments on three recent RGB-T tracking benchmarks show that our tracker 
	performs significantly better than other competing algorithms. 
\end{itemize}

\section{Related Work}
\subsection{RGB-T Tracking}

RGB-T tracking, as a branch of single target tracking, has drawn more attention in recent years~\cite{GTOT,RGBT210,CMR,DAPNet,RGBT234}. 
Existing methods~\cite{CFM,TFF,DAPNet,MANet} mainly focus on how to fuse the multimodal information for tracking. 
In~\cite{TFF}, Conaire~\emph{et al.} propose an RGB-T framework to combine the feature representation from different modalities. 
After that, some methods improve tracking accuracy using sparse coding to mine multimodal information~\cite{JSR,CMR,LGMG}. The JSR~\cite{JSR} method learns the joint sparse representation on different modalities and fuses the target likelihood using minimization operation. 
The LGMG~\cite{LGMG} tracker constructs a multi-graph descriptor to suppress the background 
effects for RGB-T tracking. The CMR~\cite{CMR} algorithm utilizes a cross-modal manifold ranking 
algorithm to address the background clutter cases in RGB-T tracking. 
Recently, deep convolutional networks have been introduced into RGB-T tracking~\cite{FANet,DAPNet,MANet} 
and have significantly improved the tracking performance. 
FANet~\cite{FANet} learns both layer-wise and modality-wise feature weights to yield discriminative features for RGB-T tracking; while DAPNet~\cite{DAPNet} applies feature fusion and pruning process to achieve more robust feature representation.
Furthermore, MANet~\cite{MANet} introduces a multi-adapter network to perform feature fusion in an end-to-end manner. 
For RGB-T tracking, it is important to exploit the information from both RGB and T modalities during the tracking process. In this work, we first develop a multimodal fusion network to conduct effective late fusion. 
Besides, we believe that motion cues are also very important for the RGB-T tracking task, and jointly 
model motion and appearance cues to improve the tracking accuracy. 

\begin{figure*}[t]
	\centering
	\includegraphics[width=1.0\linewidth]{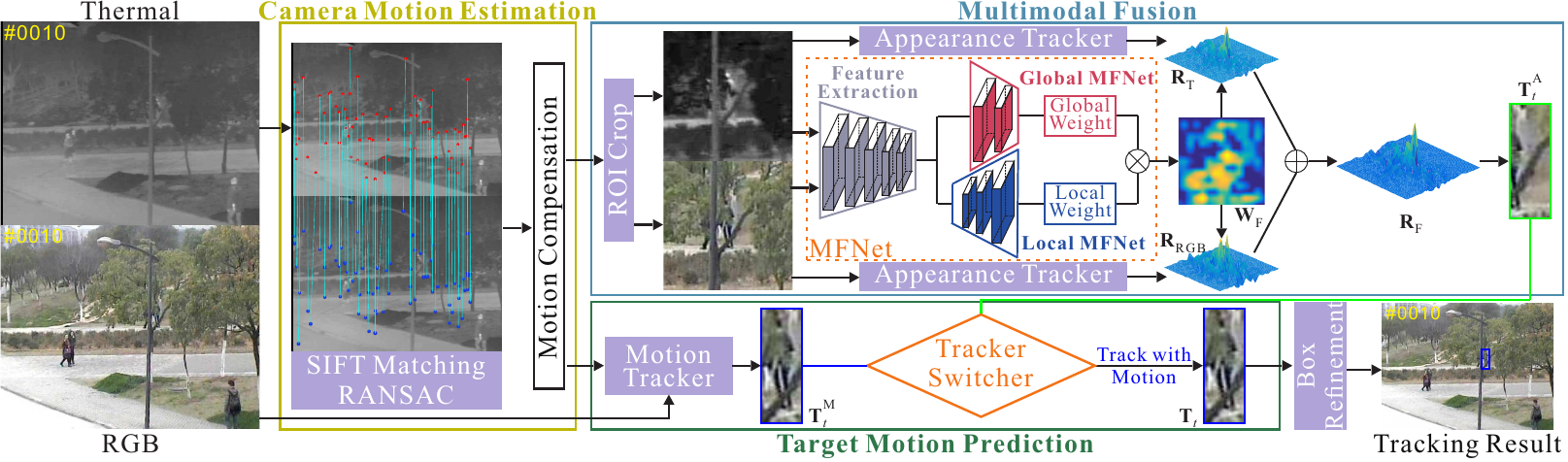}
	
	\caption{JMMAC RGB-T tracking framework. We jointly model motion and appearance cues via two main components, i.e., multimodal fusion and motion mining. Multimodal fusion aims to fuse the appearance information in two modalities and improves the tracking accuracy by our MFNet. To the specific, we model motion cues using two schemes: target motion prediction and camera motion estimation. Target motion prediction predicts target position via motion information and determines which information is more reliable for tracking by tracker switcher. Camera motion estimation attempts to compensate for camera movement, stably providing effective search regions. 
	}
	\label{fig:figframework}
\end{figure*}

\subsection{Multimodal Fusion}

Multimodal fusion attempts to integrate information from different modalities by employing the connection 
of multimodal data to obtain a more reliable classification or regression output~\cite{MIA1,AVSR1,MSA1,MSA2,AVSR2,MIA2}.
It usually can be categorized into two types: early fusion~\cite{DFF,EDFF} and late fusion~\cite{SN,QALF,OCF}. 
Early fusion, by fusing low-level feature among modalities, aims to discover the complementary information 
of different modalities. Chaib~\emph{et al.}~\cite{DFF} propose a feature fusion method for very high resolution 
remote sensing scene classification. Shao~\emph{et al.}~\cite{EDFF} conduct feature fusion for rotating machinery 
fault diagnosis via locality preserving projection. 
But the dimensionality of data expands multiply and thus leads to a high inference time. 
In contrast, late fusion combines the decisions~(e.g., classification scores or tracking responses) to obtain a 
final result via the various fusion approaches, keeping independent models to give responses for each modality 
and maintaining flexibility~\cite{MF_survey}. Zheng~\emph{et al.}~\cite{QALF} propose a late fusion method 
at the score level to evaluate the quality of features. Terrades~\emph{et al.}~\cite{OCF} combine the result of 
classifiers via non-Bayesian probabilistic framework to improve the classification performance. 
Jain~\emph{et al.}~\cite{SN} propose a score normalization approach for robust and fast late fusion. 
Recently, many researchers have applied early fusion to RGB-T tracking~\cite{DAPNet,FANet,MANet}, while the effectiveness of 
late fusion has not been exploited. 
In this work, we first attempt to introduce late fusion to the RGB-T tracking task, thereby yielding better 
performance. 

\subsection{Tracking with Motion Cues}

The motion cues (from target motion, camera motion or both) are also crucial but often ignored in designing a tracking framework.
Target motion is widely used in previous tracking frameworks, such as Kalman filter~\cite{KOT,AKF,ACDE} 
and particle filter~\cite{PF4PNT,BPF,PF-Li08,GeometricPF-Kwon}. Comaniciu~\emph{et al.}~\cite{KOT} integrate Kalman filter 
and data association for target localization and representation. Weng~\emph{et al.}~\cite{AKF} utilize the adaptive Kalman 
filter to construct a motion model for tracking in complex situations (e.g., fast motion, occlusion and illumination variation). 
Kulikov~\emph{et al.}~\cite{ACDE} propose a continuous-discrete Extended Kalman filter method for radar tracking. 
Kenji~\emph{et al.}~\cite{BPF} propose an improved particle filter method, which combines mixture particle filter and 
Adaboost. 
Kwon~\emph{et al.}~\cite{GeometricPF-Kwon} develop a geometric particle filter technique based on matrix Lie groups, 
to overcome the limitation of the traditional deterministic optimization approach. 
As for camera motion, it is difficult to model camera motion since the camera parameter is unknown for inferring 
the 3D target location~\cite{UAV70}. Furthermore, some datasets~\cite{OTB13,OTB15,GTOT} are captured 
by stationary camera or with slight camera movement, where the importance of camera motion is underestimated. 
Existing works on tracking with drones predict camera movement since the camera is far away from the target 
and the depth variation is negligible~\cite{UAV70}. 
Due to the powerful feature descriptions and learning algorithms, researchers have paid less attention to motion model when solving the tracking task~\cite{MDNet,ECO,SiamMask}. 
However, in this work, we find that the consideration of target motion and camera motion could significantly improve 
the tracking performance in RGB-T tracking task.

\section{Tracking Framework via Jointly Modeling Motion and Appearance Cues~(JMMAC)}

We propose a novel framework for robust RGB-T tracking by jointly modeling motion and appearance cues 
(JMMAC), which consists of two main components: multimodal fusion and motion mining.
The JMMAC framework is shown in Figure~\ref{fig:figframework}. To be specific, we model motion cues using 
two schemes: target motion prediction and camera motion estimation. 
The tracking process can be easily summarized as follows. First, we apply camera motion compensation to 
deal with severe camera motion. Then, we exploit the late fusion method to aggregate tracking responses from 
different modalities via the proposed MFNet. We also maintain a motion tracker. When the appearance information is unreliable, our framework can dynamically select which cue is used for target location via target motion prediction module. 
Finally, we apply the bounding box refinement to adjust the final result for better scale estimation.

\subsection{Multimodal Fusion Network for Robust Appearance Model}\label{MFNet}
The ECO~\cite{ECO} method is chosen as our base tracker due to its effectiveness. 
Two ECO trackers are created for RGB and T modalities, respectively. 
Then, the corresponding response maps are obtained in each frame, denoted as 
${\bf R}_{RGB}  \in \mathbb{R}^{M \times N}$ and ${\bf R}_{T} 
\in \mathbb{R}^{M \times N}$ (the size of search regions is $M \times N$). 
In this work, we attempt to conduct a late fusion via a linear combination manner,

\begin{equation}\label{fuse_response}
{\bf R}_{F} = {\bf W}_{F} \odot {\bf R}_{RGB} + ({\bf 1}-{\bf W}_{F}) \odot {\bf R}_{T},
\end{equation}
where ${\bf W}_{F}  \in \mathbb{R}^{M \times N}$ is a fusion weight whose elements are bounded from zero to one, and ${\bf 1}  \in \mathbb{R}^{M \times N}$ is a matrix whose elements are all one. $\odot$ denotes the elementwise production operation.
The target location can be determined based on the peak of the fused response map ${\bf R}_{F}$. 
The proposed MFNet, whose architecture is shown in Figure~\ref{fig:figframework}, aims to learn a precise pixel-wise fusion weight ${\bf W}_{F}$ for RGB-T tracking from coarse to fine.

\textbf{\emph{Note that our MFNet is effectively offline trained and directly applied for tracking without online fine-tuning.}}

\noindent{\textbf{MFNet.} Our MFNet consists of 
a shared feature extractor and two subnetworks, namely global and local MFNet. As for the feature extractor, we use the truncated VGG-M network, which is pretrained on ImageNet without fine-tuning. Our MFNet takes the fixed-size RGB and thermal patches ${\bf P}_{RGB}, {\bf P}_{T}$ as inputs, and then extracts features using the truncated VGG-M network on ${\bf P}_{RGB}$ and ${\bf P}_{T}$ to obtain high-level features~(Conv-5) ${\bf F}_{RGB}$ and ${\bf F}_{T}$. Then, we concatenate them and send them to the subnetworks: (a) global MFNet, outputting a global weight ${w}_{\rm G} \in {\mathbb  R}^1$ to emphasize the contributions of different modalities; and (b) local MFNet, providing a pixel-level local weight ${\bf W}_{L} \in {\mathbb  R}^{M \times N}$ to consider the distractors within each modality. 
The final weight ${\bf W}_{F} \in {\mathbb  R}^{M \times N}$ can be obtained by 
\begin{equation}
{\bf W}_{F} = {w}_{G} * {\bf W}_{L},
\end{equation}
where ${w}_{G}$ and ${\bf W}_{L}$ are constrained to $(0,1)$ by using a Sigmoid layer. 
Figure~\ref{fig:MFNet} provides a visual example of the results obtained by our MFNet, indicating that the global weight $w_{G}$ depicts the importance of different modalities and the local weight ${\bf W}_{L}$ suppress the influence of distractors within each modality during the tracking process.

\textbf{(1) Global MFNet}. We first exploit the complementarity of RGB and T modalities by using a global MFNet to obtain the weight over the whole context. 
The output ${w}_{G}$ of our global MFNet reflects the importance of each modality during the tracking process (see Figure~\ref{fig:MFNet}).
Our global MFNet contains two convolution layers, whose filter size is 3$\times$3$\times$256 and 9$\times$9$\times$1. Each layer is followed by rectified linear unit (ReLU) and local response normalization~(LRN).
	
\textbf{(2) Local MFNet}. Though cross-modality divergence has been considered by our global MFNet, distractors in each individual modality is also harmful to robust tracking. 
Thus, we introduce a local MFNet to suppress the influence of distractors and achieve more accurate response maps. 
Our local MFNet acts as an attention mechanism to obtain a finer weight map, and is constructed based on the U-Net architecture~\cite{UNet}. 
Specifically, our local MFNet consists of two deconv layers where the kernel size is 3 $\times$ 3 $\times$ 256 and 3 $\times$ 3 $\times$ 1 and a bilinear sampling layer which is adopted to resize the weight map to fit the size of response map. It is noted that each of deconv layers is followed by ReLU. In Figure~\ref{fig:MFNet}, as a supplement to global MFNet(${\bf R}_{G}$ denotes the response only using global MFNet), local MFNet can suppress the distractors and obtains accurate response with high peak-to-sidelobe rate~(PSR).
	
\textbf{(3) Loss function}.
Similar to the classical correlation-filter~(CF)-based methods~\cite{DSST_PAMI,KCF}, we learn our MFNet by minimizing the squared Euclidean distance between the desired response $\bf Y$ and the fused response map ${\bf R}_{F}$, 
\begin{equation}
{\cal L} = ||{\bf R}_{F}-{\bf Y}||_2^2	
\end{equation}
where the ground truth label $\bf Y$ is a 2-D Gaussian map whose peak denotes the target location. 

\begin{figure}[t]
	\centering
	\includegraphics[width=1.0\linewidth]{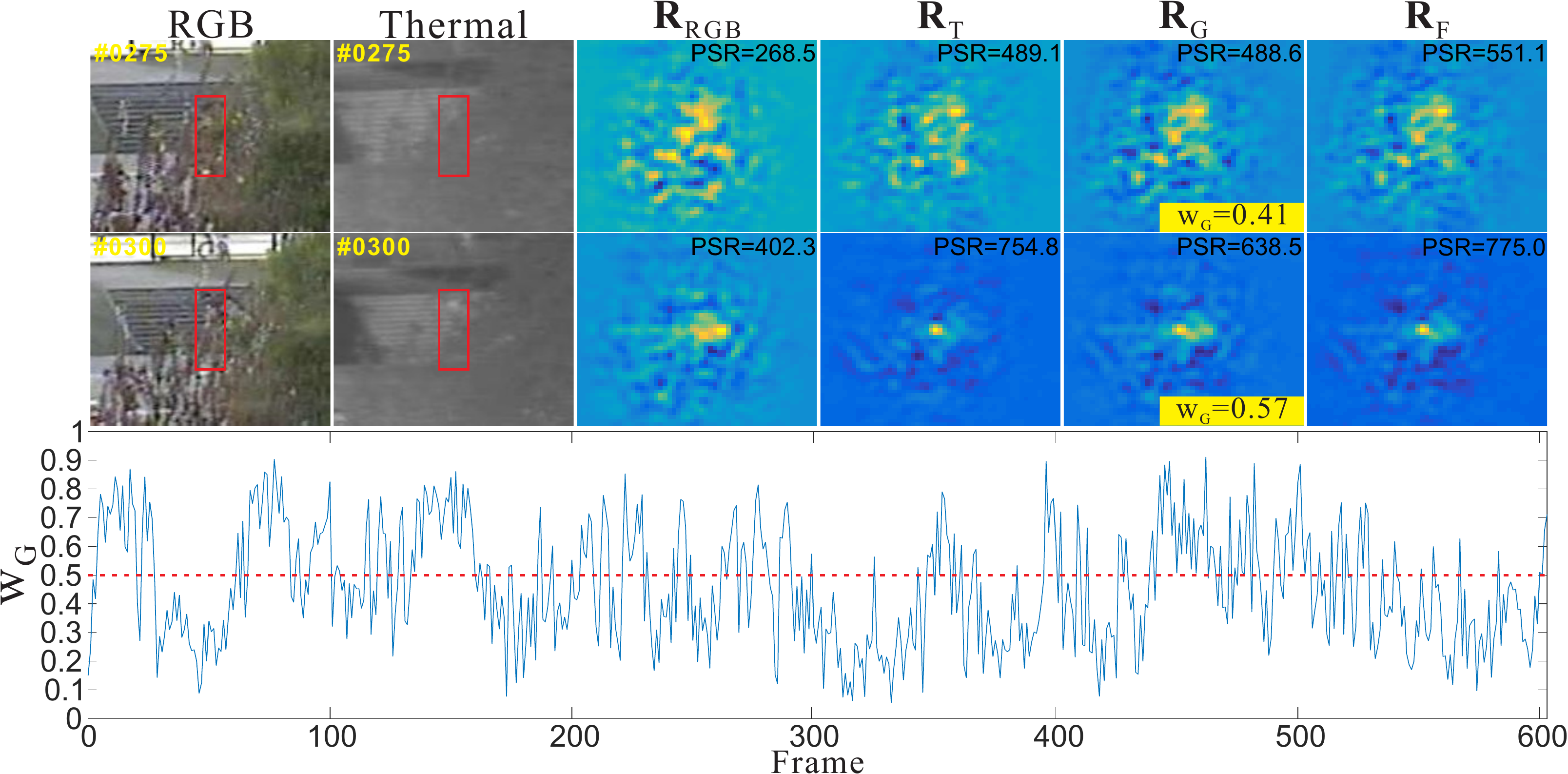}
	\caption{Qualitative results of our MFNet. The global weight $w_{G}$ dynamically varies to consider the contributions of two modalities during tracking process. The local weight ${\bf W}_{L}$ provides a necessary supplement, resulting in a robust response map. The visualization of ${\bf W}_{L}$ can be found in supplementary video.}
	\label{fig:MFNet}
\end{figure}

\subsection{Motion Modeling for Robust RGB-T Tracking}
As presented in Section~\ref{sec1} and Figure~\ref{fig:fig1}, motion information is also very important for RGB-T tracking especially when the appearance information is unreliable due to complex variations (e.g., extreme illumination, low resolution, camera motion and occlusion). 
However, few existing RGB-T trackers have paid attention to mining motion cues. 
In this work, we explicitly divide motion information into target motion and camera motion, and design two different modules to handle them, namely, target motion prediction~(TMP) and camera motion estimation~(CME). 

\begin{figure}[t]
	\centering
	\includegraphics[width=1.0\linewidth]{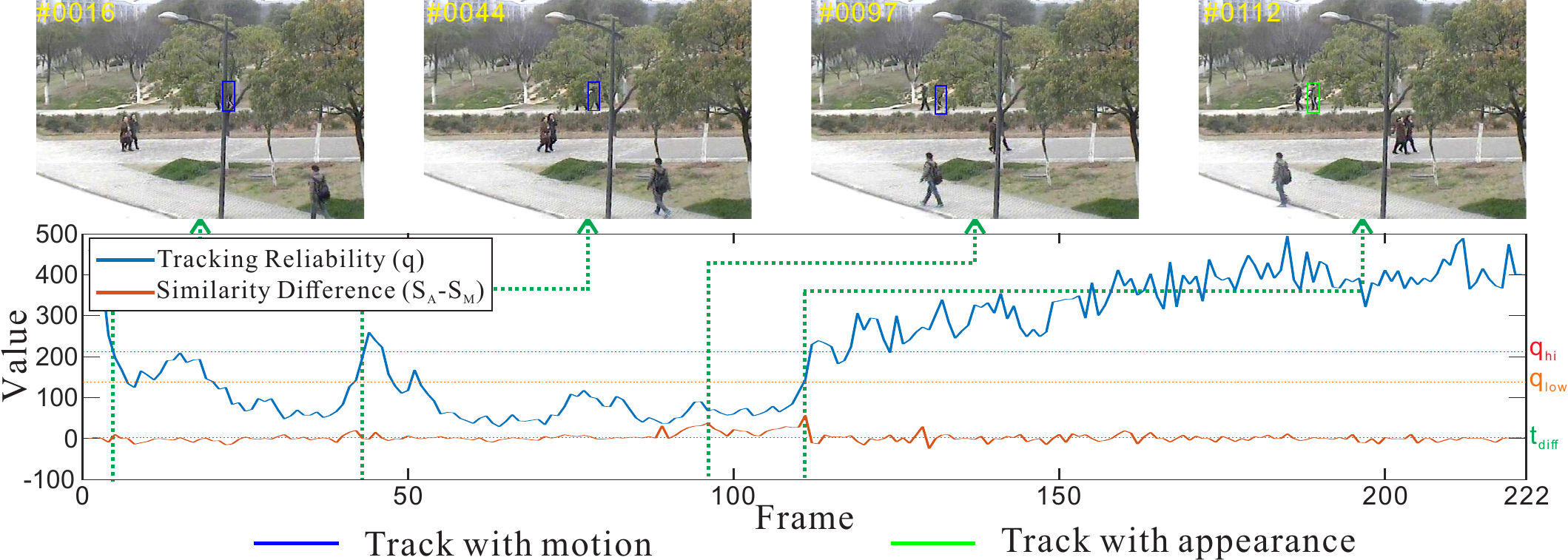}
	\caption{Effectiveness of our TMP scheme. This figure indicates our TMP module works well in switching between appearance and motion trackers by combining MAX-PSR and template matching. }
	\label{fig:occlusion_analysis}
\end{figure}
	
\subsubsection{Target Motion Prediction}\label{OCC}
Appearance tracker can not give accurate results when the appearance information is unreliable especially in low resolution and occlusion cases. 
In this situation, we design the target motion prediction scheme by introducing a motion tracker to predict target motion and proposing a tracker switcher to dynamic select which tracker is more reliable as final result.

\noindent{\textbf{Switcher.} In our framework, we jointly model the appearance and motion cues by maintaining the appearance and motion trackers. We argue that in most cases, appearance is much more discriminative than motion information, while appearance can hardly help locate the target when the target is occluded or in low resolution. Thus, we design a simple yet efficient switching mechanism to determine which cue is more suitable for tracking, which simultaneously consider the reliability of response map and similarities between the target template and tracking results. 
		
\textbf{(1) MAX-PSR.} First, we evaluate the reliability of the appearance tracker with a self-adaptive method presented in~\cite{FCLT}. This method combines the PSR and maximum value of the response map $\bf R$. The reliability value $q$ is defined as,

\begin{equation}\label{eq_TrackingQuality}
	q = {\rm PSR}({\bf R}) \times {\rm max}({\bf R}), 
\end{equation}
\vspace{-4mm}
\begin{equation}
	{\rm PSR} = \frac{{\rm max}({\bf R})-{\rm mean}({\bf R})}{{\rm var}({\bf R})}
\end{equation}
and ${\rm max}(\bf R), {\rm mean}(\bf R),{\rm var}({\bf R})$ denote the maximum value, mean value and variance of the response map, respectively. 
The larger MAX-PSR is, the more reliable results we contain. 
However, as shown in Figure~\ref{fig:occlusion_analysis}, the $q$ value cannot always reflect the tracking reliability. 
Besides, online updating with noisy samples makes the response not sensitive to appearance variation. 
Thus, we develop the MAX-PSR method by considering the template similarity with template matching method to ensure a stable switching. The template matching scheme is not affected by noisy observations since it is without online update. 

\textbf{(2) Template Matching.} We also compare the results of both appearance and motion trackers using the template matching method in RGB modality. 
First, we use ${\bf T}_{1}$ to denote the template image of target in the first frame. ${\bf T}^{A}_t$ and ${\bf T}^{M}_t$ are target regions obtained from appearance and motion trackers in the $t$-th frame.
Then, we define two similarity scores $s_{A}$ and $s_{M}$ to evaluate the reliability of tracking results from appearance and motion trackers, respectively. 
\begin{equation}\label{simi1}
	s_{A} = {\cal TM}({\bf T}_{1},{\bf T}^{A}_t), 
\end{equation}
\vspace{-4mm}
\begin{equation}\label{simi2}
	s_{M} = {\cal TM}({\bf T}_{1},{\bf T}^{M}_t), 
\end{equation}
where ${\cal TM}(\cdot)$ denotes the template matching function. To be specific, we use the deformable diversity similarity (DDIS)~\cite{DDIS} method to conduct matching and calculate the similarity scores. 
The DDIS method measures the similarity between the target and template images based on the diversity of feature matches, and therefore is robust to complex deformation, background clutter and occlusion. 
More implementation details can be found in~\cite{DDIS}. The tracker switcher considers both offline information, i.e., the initial target template, and online response map, thus achieving good switch between the appearance and motion trackers. Figure~\ref{fig:occlusion_analysis}  shows that our tracker switcher can select the meaningful information, thereby yielding a robust tracking result. The overall TMP scheme is summarized in Algorithm~\ref{Algorithm1}.

\subsubsection{Camera Motion Estimation}

\begin{figure}[t]
	\centering
	\includegraphics[width=1.0\linewidth]{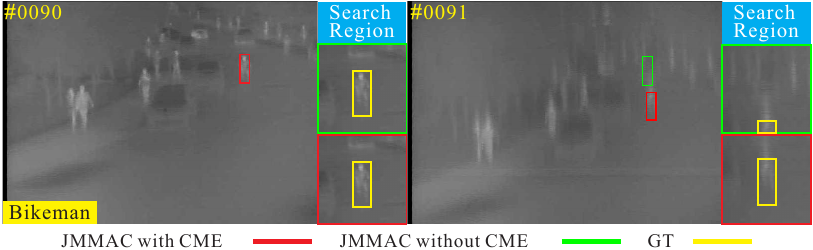}
	\caption{Visual results of our tracker with and without CME.}
	\label{fig:cmanalysis}
\end{figure}
Since RGB-T images are often captured by the high-altitude cameras being far from the targets, we assume that both target movement and depth variation of the target are very small. 
Thus, we directly model the camera motion in the 2D image plane rather than apply depth prediction. 
To be specific, we estimate camera motion by calculating the transformation matrix $\bf O$ between the reference image ${\bf I}_r(x,y)$ and search image ${\bf I}_s(x,y)$ in thermal modality. 
The transformation matrix can be obtained by 
\begin{equation}\label{Transformation}
	(x',y') = {\cal T}(x,y; \bf O),
\end{equation}
where $(x',y')$ are coordinates of the key point in the search image corresponding to $(x,y)$ in the reference image and ${\cal T(\cdot)}$ is the transformation function with the parameter matrix 
\textbf{O}. 
The affine transformation with six parameters is adopted in default. 
First, key points of both reference and search images are extracted with the scale-invariant feature transform (SIFT) method~\cite{SIFT}.   
Then, the M-estimator sample consensus algorithm is used to match key points and exclude outliers.
The obtained transformation matrix $\bf O$ can compensate for the effect of camera motion. 
Figure~\ref{fig:cmanalysis} provides an example with large camera motion, indicating our CME scheme facilitates obtaining a stable search region.

\subsection{Tracking with JMMAC}
The overall tracking framework has been presented in Figure~\ref{fig:figframework}, some additional explanations are as follows. 
		
\noindent{\textbf{Model Updating Scheme.}
Traditional CF-based methods update the filter every frame~\cite{KCF,DSST_PAMI} or fixed interval~\cite{ECO} with the fixed~\cite{KCF} or adaptive learning rate~\cite{CCOT}. 
However, with such updating scheme, the filter may be degraded by the corrupted samples when the appearance information is unreliable. 
Thus, we skip the filter updating process when the motion tracker is applied to track the target. 
We update the motion tracker every frame to record the target trajectory generated by the appearance tracker. 
If the motion cues are utilized for tracking, the motion tracker predicts the target's position and the tracker is updated with the predicted result.
			
\noindent{\textbf{Motion Tracker.}} We apply Kalman filter tracker as our motion tracker.
To make the paper self-contained, we detail the Kalman filter. 
We utilize Kalman filter to predict the target location by motion cues, i.e., position and velocity. In our method, we assume that target maintains constant velocity during sampling. 
In $t$-th frame, Kalman filter aims to estimate the target state ${\bf x}_{t} = (p_x,v_x,p_y,v_y)^{\rm T}$ via a linear difference equation, 
\begin{equation}
{\bf x}_{t} = {\bf A}{\bf x}_{t-1} + {\bf w}_{t-1},
\end{equation}
where $(p_x,p_y)$ denotes the coordinate of target center, $(v_x,v_y)$ is the target velocity in both X-axis and Y-axis directions. ${\bf A}$ is state transformation matrix which is defined as follow,
\begin{equation}
{\bf A} = \begin{bmatrix}
1 & 1 & 0 & 0 \\ 
0 & 1 & 0 & 0 \\ 
0 & 0 & 1 & 1 \\ 
0 & 0 & 0 & 1
\end{bmatrix}.
\end{equation}
${\bf w}_{t-1}$ denotes the process noise with normal probability distribution, i.e., ${\bf w}_{t-1} \sim N(0,\bf Q)$, where $\bf Q$ denotes the noise covariance matrix.

Two steps are included in Kalman filter tracking process, i.e., prediction and updating.
In the prediction step, the target position is obtained via dynamic model expressed as, 
\begin{equation}
{\bf \hat z}_t = {\bf H}{\bf x}_{t}+{\bf v}_{t}.
\end{equation}

where ${\bf x}_{t}$ and ${\bf \hat z}_t$ denote the target state and predicted measurement in the current frame, respectively. ${\bf H} \in \mathbb{R}^{2 \times 4}$ is measurement matrix which is defined as, 
\begin{equation}
{\bf H} = \begin{bmatrix}
1 & 0 & 0 & 0 \\ 
0 & 0 & 1 & 0 \\ 
\end{bmatrix},
\end{equation}
and ${\bf v}_{t-1}$ is the measurement noise and ${\bf v}_{t-1} \sim N(0,\bf R)$. $\bf R$ is the measurement covariance matrix.
In the updating step, the target state is updated with the actual measurement ${\bf z}_t$. More details can be found in \cite{introKF}. In practice, the noise covariance $\bf Q$ and measurement covariance $\bf R$ are set to, 
\begin{equation}
{\bf Q} = \begin{bmatrix}
25 & 0 & 0 & 0 \\ 
0 & 10 & 0 & 0 \\ 
0 & 0 & 25 & 0 \\ 
0 & 0 & 0 & 10 \\ 
\end{bmatrix},
\end{equation}\begin{equation}
{\bf R} = \begin{bmatrix}
25 & 0 \\ 
0 & 25 \\ 
\end{bmatrix}.
\end{equation}
When the tracking result of Kalman filter is chosen as final result, Kalman filter is not updated since the actual measurement is unavailable. Otherwise, we consider the tracking result of appearance tracker as the actual measurement used for updating Kalman filter.

\noindent{\textbf{Bounding Box Refinement.}
The existing CF-based trackers estimate scale variation by sampling the search region among multiple resolutions 
and finding the corresponding scale with the maximum response. 
However, in this manner, the bounding box may not tightly capture the target since the ratio variation is ignored. 
In this work, we introduce a simple box refinement process followed by scale estimation with a real-time YOLOv2~\cite{YOLO} detector to alleviate this problem. 
The detector, which is pretrained on COCO~\cite{COCO} dataset, is efficiently applied in a small region surrounding the target without utilizing the category label. 
Note that, the bounding box refinement is performed on the visible modality and is disabled in some extreme cases (e.g., illumination influence and low resolution).
				
\begin{algorithm}[t]
	\caption{Target motion prediction (TMP).}
	\label{Algorithm1}
	\KwIn{Fused response map ${\bf R}_{F}$, tracking results ${\bf T}^{A}_t$ and ${\bf T}^{M}_t$, and the target template ${\bf T}_{1}$.}
	\KwOut{ Final tracking result ${\bf T}_t$}
		{	
		    Calculate the tracking reliability $q$ via equation~(\ref{eq_TrackingQuality})\\
			Obtain the similarities $s_{A}$ and $s_{M}$ via equations~(\ref{simi1}) and~(\ref{simi2}).\\
			\If{($q>{q_{hi}}$ and $s_{A}>{s_{hi}}$) or ($q>{q_{low}}$ and $s_{A}>{s_{low}}$ 
			and $(s_{A}-s_{M}) > t_{diff}$ ) or $max(s_{A},s_{M})< t_{disable}$}
			{${\bf T}_t$ $\leftarrow$ ${\bf T}^{A}_t$ (using the appearance tracker)\;
			\lElse{${\bf T}_t$ $\leftarrow$ ${\bf T}^{M}_t$ (using the motion tracker)}
		}
	}
\end{algorithm}
\noindent{\textbf{Implementation Details.}
We adopt a popular CF-based tracker, ECO~\cite{ECO}, as our baseline. Both deep features~(Conv1 and 
Conv5 of VGG-M for visible modality, Conv1, Conv4 and Conv5 of VGG-M for thermal modality) and 
handcraft features~(HOG~\cite{HOG} and Color Name~\cite{CN}) are used as feature representation. 
For training MFNet, we first train the global MFNet and freeze it, and then train the local MFNet. 
After that, we jointly fine-tune the overall MFNet. 
We random select training pairs~(Two frames are included in a training pair. One is used for tracker initialization, another is used for learning MFNet.) within an interval of 5 frames, and then crop the ROI and obtain the 
training patches ${\bf P}_{RGB}$ and ${\bf P}_{T}$. 
The resolutions of ${\bf P}_{RGB}$ and ${\bf P}_{T}$ are all set to 200 $\times$ 200. 
The learning rate is set to $1e^{-5}$ when both global and local MFNet are trained and is set to $1e^{-7}$ in 
fine-tuning the overall MFNet, and the batch size is chosen as 8.
The weight decay and momentum are set to 0.0005 and 0.9, respectively. 
We train our MFNet network using VOT19-RGBT when conducting evaluation on GTOT, and train 
it using GTOT when testing trackers on VOT19-RGBT and RGBT234. 
When drastic camera motion is detected, the multimodal fusion, target motion prediction and box refinement operations are suspended since the appearance information is unreliable. 
To further alleviate the computation brought by CME and TMP, we apply a series of pre-processing operations. The CME pre-processing operation based on frame difference calculates the number of pixels with large variation between current and previous frames. Then, the frames with slight change are ignored to conduct CME module. Before applying template matching in TMP, the reliability value $q$ is measured. If $q$ is higher than 250, which indicates the appearance tracker is much more reliable for tracking, we suspend the template matching processing to boost the tracker.
In target motion prediction module, the DDIS method extracts the deep features with the VGG-19 Network in default as described in~\cite{DDIS}. 
The $q_{hi}$ and $s_{hi}$ are set to 210 and 15, $q_{low}$ and $s_{low}$ are set to 135 and 17, and $t_{diff}$ is 3, respectively.
\section{Experiments}
\label{experiment}
We evaluate our tracker in comparison with other competing ones using three RGB-T tracking datasets (GTOT~\cite{GTOT}, RGBT234~\cite{RGBT234}, and VOT19-RGBT~\cite{VOT2019}). 
Our tracker (referred as {\bf JMMAC}) is implemented by MATLAB 2015b, Intel-i9 CPU with 64G RAM and a RTX2080Ti GPU with 11G memory.
\emph{Both training and testing codes will be publicly available.}

\noindent{\textbf{Datasets and Metrics.}}
The GTOT dataset~\cite{GTOT}, constructed in 2016, contains 50 grayscale-thermal sequences 
annotated with seven challenging attributes, including occlusion~(OCC), large-scale variation~(LSV), 
fast motion~(FM), low illumination~(LI), thermal crossover~(TC), small object~(SO), and deformation~(DEF).  
The RGBT234 dataset~\cite{RGBT234}, proposed in 2019, is the largest RGB-T tracking 
benchmark at present. This dataset includes 234 sequences with more than 234,000 frames and extends 
the number of attributes to 12. These attributes include no occlusion~(NO), partial occlusion~(PO), 
heavy occlusion~(HO), low illumination~(LI), low resolution~(LR), thermal crossover~(TC), 
deformation~(DEF), fast motion~(FM), scale variation~(SV), motion blur~(MB), camera moving~(CM) and background clutter~(BC). 
The comparisons follow the one pass evaluation (OPE) rule with Success Rate (SR) and Precision Rate (PR), which are widely used in~\cite{OTB13,OTB15}. 
Given that two modalities exist in the RGB-T tracking task, the SR and PR values are used to measure the 
maximum Intersection over Union~(IoU) and the minimum center location error between two modalities frame by frame (denoted as MSR and MPR~\cite{GTOT,RGBT234}).
The VOT19-RGBT~\cite{VOT2019} challenge selects 60 representative video clips from 
RGBT234~\cite{RGBT234} and adopts the Expected Average Overlap~(EAO) rule (along 
with tracking accuracy~(A) and robustness~(R)) to emphasize the short-term evaluation. 
Besides, the EAO values are calculated based on the trackers' results and the ground 
truths of the thermal modality (we refer the readers to~\cite{VOT2019} for more details).  
				
\noindent{\textbf{Compared Algorithms.}}
By using GTOT~\cite{GTOT} and RGBT234~\cite{RGBT234}, we compare our tracker with 
eight competing RGB-T methods, including MANet~\cite{MANet}, FANet~\cite{FANet}, 
ECO+RGBT, TODA~\cite{TODA}, DAPNet~\cite{DAPNet}, SiamFC+RGBT, SGT~\cite{RGBT210}, and 
CMR~\cite{CMR}\footnote{ECO+RGBT and SiamFC+RGBT are our implemented trackers, which 
	improve the traditional ECO~\cite{ECO} and SiamFC~\cite{SiamFC} ones by concatenating RGB 
	and thermal features as input features.}. 
All these trackers have achieved top-ranked performance or serve as baselines in both GTOT and 
RGBT234 benchmarks and they can be categorized into deep-learning-based (MANet, SiamFC+RGBT, 
TODA, ECO+RGBT, DAPNet and FANet), CF-based (ECO+RGBT), 
graph-learning-based (SGT) and sparse-coding-based (CMR) ones. 
In addition, we compare our JMMAC method with eight recent trackers in the official 
VOT2019~\cite{VOT2019} challenge report.

\subsection{Quantitative Evaluation}

\noindent{\textbf{GTOT~\cite{GTOT}.}} First, we compare our tracker with other methods using the GTOT dataset and report both success and precision plots in Figure~\ref{fig-gtot-sota}. 
We can see that our JMMAC method obtains the best performance with 73.2\% 
and 90.1\% in success and precision scores, respectively. 
Compared with the most recent tracker (also the second best one), MANet, our algorithm 
achieves 0.8\% improvement in success and 0.7\% in precision. 
As shown in Table \ref{GTOT_attribute}, the attribute-based comparison also shows the capability in handling occlusion~(OCC), large scale variation~(LSV), fast motion~(FM), low illumination~(LI), thermal crossover~(TC), small object~(SO) and deformation~(DEF).
\begin{table*}[h]
	\caption{Attribute-based comparison with 8 state-of-the-art trackers on the GTOT dataset. Maximum Success Rate and Maximum Precision Rate~(MSR$\backslash$MPR \%) are used for evaluation. It is indicated that JMMAC outperforms all the competitors with a large margin.}\label{GTOT_attribute}
	\footnotesize
	\begin{center}
		\begin{tabular}{cccccccc|c}
			\hline
			& OCC & LSV & FM & LI & TC & SO & DEF & ALL\\
			\hline
			MANet & 69.6/{\color{red} \bf88.2} & 70.1/87.6 & 69.9/87.6 & 71.2/89.0 & 71.0/89.0 & 70.8/89.7 & 71.5/90.1 & 72.4/89.4\\
			FANet & {\color{red} \bf 70.3}/86.4 & 69.2/84.0 & 68.4/83.2 & 70.2/89.9 & 70.8/86.7 & 70.8/88.1 & 71.8/89.1 & 72.8/89.1\\
			ECO+RGBT &66.4/81.1  &69.3/83.7 & 70.0/83.5& 70.7/85.4& 69.6/84.2 &69.3/85.0 &69.0/85.0  &67.7/82.1 \\
			TODA &63.5/84.6 &64.4/84.8 & 64.2/84.8& 66.3/85.1&66.0/85.3 &66.3/86.7  &67.6/86.9 &67.7/84.3\\
			DAPNet & 67.4/87.3&66.1/86.0 &65.3/85.2 &67.7/86.9 &68.0/87.5 & 68.2/88.6&69.6/89.1 &70.7/88.2\\
			SiamFC+RGBT &59.3/74.7 & 62.2/77.7&61.1/75.8  & 60.5/74.7&61.1/76.0 &60.3/76.1 & 60.1/75.2&60.6/74.1 \\
			SGT &56.7/81.0 &55.7/82.6  &55.7/82.0 &59.0/84.3 &59.6/84.4 &60.0/85.7 & 62.1/86.7&62.8/85.1 \\
			CMR &62.6/82.5 &64.7/83.9 &64.7/83.8 & 65.8/85.5 &64.9/84.4 &64.2/84.8 &64.4/84.8 &64.3/82.7\\
			\bf JMMAC & 68.4/84.0 &{\color{red} \bf71.5/87.4} &{\color{red} \bf72.4/87.9} &{\color{red} \bf73.9/90.3} & {\color{red} \bf73.0/89.9}&{\color{red} \bf 73.2/90.7} &{\color{red} \bf73.6/91.7} &{\color{red} \bf73.2/90.1}\\
			\hline
		\end{tabular}
	\end{center}
\end{table*}

\begin{figure}[!t]
	\centering
	\includegraphics[width=1.0\linewidth]{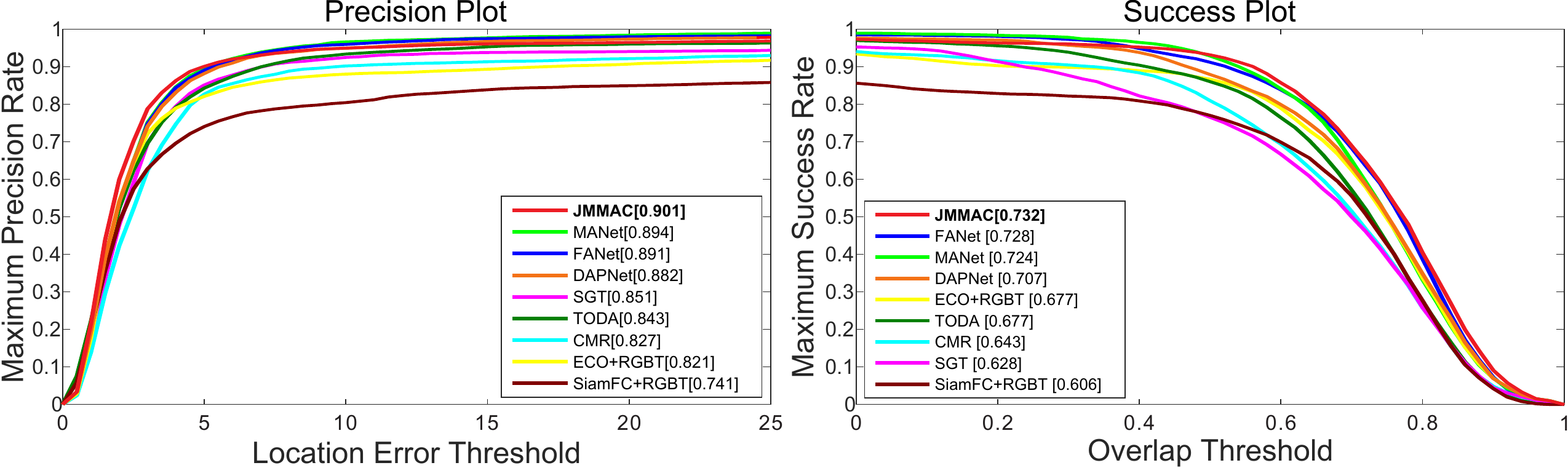}
	\caption{Performance evaluation on the GTOT in terms of success and precision plots.}
	\label{fig-gtot-sota}
\end{figure}

\begin{figure}[!t]
	\centering
	\includegraphics[width=1.0\linewidth]{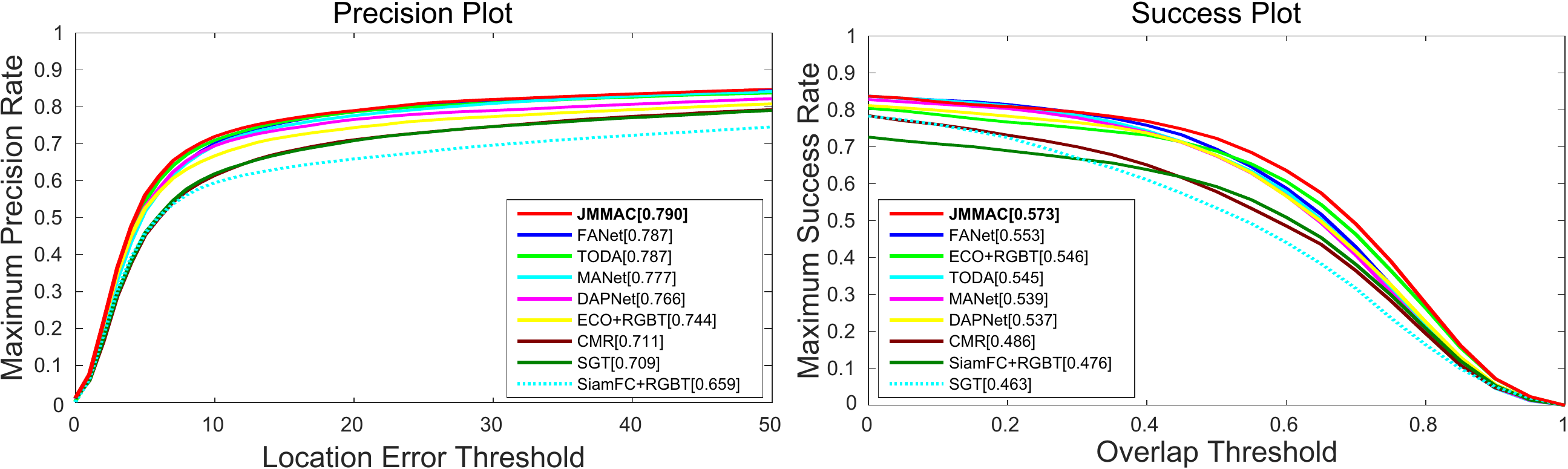}
	\caption{Performance evaluation on RGBT234 in terms
		of success and precision plots.}
	\label{fig-rgbt234-sota}
\end{figure}
\noindent{\textbf{RGBT234~\cite{RGBT234}.}} Second, we evaluate JMMAC 
method using the large-scale RGBT234 dataset and report the related results in 
Figure~\ref{fig-rgbt234-sota}. Overall, our tracker is superior to all the compared algorithms. 
Table~\ref{RGBT234} summarizes the JMMAC's performance in handling different challenging 
factors. Our tracker works very well in dealing with occlusion, low illumination, deformation, scale variation, 
motion blur, and camera moving. 
Those challenges result in unreliable appearance information and therefore renders the tracker 
easy to drift. 
An important reason is that our method develops an effective multimodal fusion network to 
learn a robust appearance model.
Additionally, our camera motion compensation and target motion prediction schemes further 
alleviate the effects of unreliable appearance information. 

\begin{table*}[!t]
	\caption{Attribute-based evaluation with nine state-of-the-art trackers on the RGBT234 dataset. Maximum success rate and maximum precision rate~(MSR/MPR) are used for evaluation. The top performance is marked in {\color{red}\bf red} font.}\label{RGBT234}
	\footnotesize
	\begin{center}
		\begin{tabular}{cccccccccc}
			\hline
			& MANet & FANet & ECO+RGBT & TODA & DAPNet & SiamFC+RGBT & SGT & CMR & {\bf JMMAC} \\
			\hline
			NO & 64.6/88.7 & 65.7/88.2 & 65.7/88.5& 64.6/89.3& 64.4/90.0& 61.2/81.6& 55.9/86.8& 61.6/89.5&  {\color{red}\bf 69.4/93.2}\\
			PO &56.6/81.6 & 60.2/{\color{red}\bf86.6} & 58.8/80.3& 57.2/82.7& 57.4/82.1& 49.4/69.3&49.0/74.8& 53.6/77.7 &  {\color{red}\bf61.1}/84.1\\
			HO & 46.5/68.9 & 45.8/66.5 & 45.4/62.2& 47.4/{\color{red}\bf69.8}& 45.7/66.0& 39.7/55.6& 39.2/59.9& 37.7/56.3& {\color{red}\bf48.3}/67.7\\
			LI & 51.3/76.9 & 54.8/80.3 & 58.8/82.9& 55.3/80.3& 53.0/77.5&47.7/67.4 & 44.4/68.7& 49.8/74.2&  {\color{red}\bf58.8/84.0}\\
			LR & 51.5/75.7 & {\color{red}\bf53.2/79.5} & 48.6/71.5& 52.2/78.4&51.0/75.0& 42.5/62.1 & 48.0/75.6& 42.0/68.7& 51.7/77.1\\
			TC & 54.3/75.4 & 54.9/76.6 & {\color{red}\bf57.6/77.9}& 50.7/74.0& 54.3/76.8& 44.4/60.8& 45.3/72.7& 44.3/67.5& 52.6/74.9\\
			DEF &52.4/72.0 & 52.6/72.2 & 49.8/66.3& 51.6/{\color{red}\bf74.3}& 51.8/71.7& 46.1/62.4& 46.6/67.7& 47.3/66.7& {\color{red}\bf52.9}/70.6\\
			FM & 44.9/69.4 &43.6/68.1 & 43.9/63.8& {\color{red}\bf48.0/75.3}& 44.3/67.0& 38.6/58.0&39.0/66.6& 38.4/61.3 &  41.7/61.0\\
			SV & 54.2/77.7 & 56.3/78.5 & 57.4/76.2& 55.4/79.2 & 54.2/78.0& 48.9/67.1& 43.3/69.3& 49.3/71.0& {\color{red}\bf61.6/83.7}\\
			MB & 51.6/72.6 & 50.3/70.0 & 49.9/67.0& 50.1/70.7& 46.7/65.3& 39.2/52.9& 42.0/62.2& 42.7/60.0&  {\color{red}\bf54.9/75.1}\\
			CM & 50.8/71.9 & 52.3/72.4 & 51.3/68.9& 49.3/69.8& 47.4/66.8& 43.6/59.6& 43.8/64.8& 44.7/62.9& {\color{red}\bf55.6/76.2}\\
			BC & 48.6/73.9 & 50.2/75.7 & 47.3/67.9& {\color{red}\bf51.3/77.1}& 48.4/71.7& 41.0/57.9& 40.4/63.9&39.8/63.1& 48.5/68.7\\
			\hline
			\bf ALL & 53.9/77.7 & 55.3/78.7 & 54.6/74.4&54.5/78.7& 53.7/76.6& 47.6/65.9 & 46.3/70.9& 48.6/71.1& {\color{red}\bf57.3/79.0}\\
			\hline
		\end{tabular}
	\end{center}
\end{table*}
				
\noindent{\textbf{VOT19-RGBT~\cite{VOT2019}.}}
Finally, we test our tracker using the VOT19-RGBT dataset and report EAO, A and 
R values in Table~\ref{tab-vot19-rgbt}. The results of other competing methods are 
obtained from the official VOT2019~\cite{VOT2019} challenge report. Our 
JMMAC tracker performs the best for all three metrics with significant performance superiority. Compared with the second-rank tracker~(SiamDW\_T), We achieve 26.8\% relative improvement in EAO with more accurate results and less failure times.

\begin{table*}[t]
	\caption{Comparison results on VOT19-RGBT. JMMAC outperforms all the competitors by a large margin. The top three results are in {\color{red}\bf RED}, {\color{blue}blue} and {\color{green}green} fonts.} 
	\label{tab-vot19-rgbt}
	\scriptsize
	\begin{center}
		\begin{tabular}{ccccccccc}
			\hline
			Trackers & GESBTT & CISRDCF & MPAT & MANet & FSRPN & mfDiMP & SiamDW\_T & \bf JMMAC \\
			A~($\uparrow$) & \color{green}0.6163 & 0.5215 & 0.5723 & 0.5823 & \color{blue}0.6362 & 0.6019 & 0.6158 & \color{red}\bf0.6597 \\
			R~($\uparrow$) & 0.6350 & 0.6904 & 0.7242 & 0.7010 & 0.7069 & \color{blue}0.8036 & \color{green}0.7839 & \color{red}\bf0.8235\\
			EAO~($\uparrow$) & 0.2896 & 0.2923 & 0.3180 & 0.3463 & 0.3553 & \color{green}0.3879 & \color{blue}0.3925 & \color{red}\bf0.4978 \\
			\hline
		\end{tabular}
	\end{center}
	
\end{table*}

\subsection{Qualitative Evaluation}\label{QR}
Figure~\ref{fig:qualitive-analysis} illustrates some qualitative results from VOT19-RGBT to compare our JMMAC and 
with other trackers in handling several challenging cases, such as camera motion 
(\emph{Baby}), low resolution (\emph{Car37}), scale variation (\emph{Caraftertree}), 
and occlusion (\emph{Greyman}).
Our JMMAC tracker, which utilizes both appearance and motion cues, can  
capture the tracked object accurately in camera motion, low resolution and occlusion cases.
And, JMMAC with box refinement module can obtain a more tight bounding box in sequence `Caraftertree', where the object has large size variation.
\begin{figure}[t]
	\centering
	\includegraphics[width=1.0\linewidth]{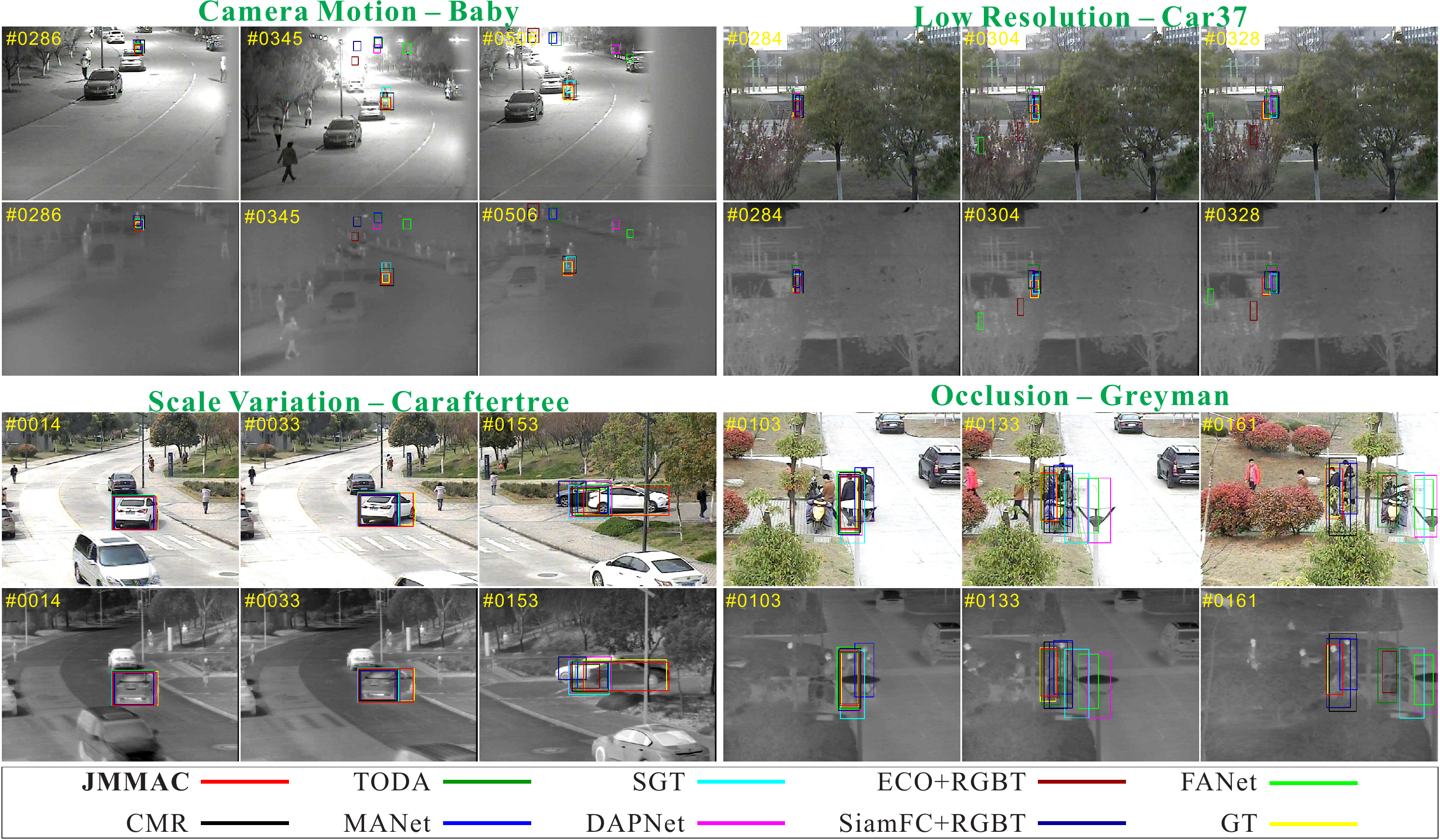}
	\caption{Representative visual results of our JMMAC and other state-of-the-art trackers 
		on the VOT-RGBT dataset. }
	\label{fig:qualitive-analysis}
	\vspace{-4mm}
\end{figure}

\subsection{Ablation Analysis}\label{Ablation_study}
	\begin{table*}[!t]
		\caption{Quantitative results of different fusion methods on the image fusion task using the dataset 
			presented in~\cite{FusionGAN}. Our MFNet achieves the competitive performance with 
			real-time speed.}\label{table_IF}
		\scriptsize
		\begin{center}
			\begin{tabular}{ccccccccccc}
				\hline
				Method & EN ($\uparrow$) & MI ($\uparrow$) & Qabf ($\uparrow$) & $\rm FMI_{w}$ ($\uparrow$)& Nabf~($\downarrow$) & SSIM ($\uparrow$) & MS\_SSIM ($\uparrow$)& FPS ($\uparrow$)\\
				\hline
				DenseFuse~\cite{DenseFuse} & 6.841 & 13.683 &0.448  & \color{red}\bf 0.432 & 0.081 & 0.710 &\color{red}\bf 0.932 &1.8 \\
				FusionGAN~\cite{FusionGAN} & 6.572 & 13.144 & 0.234& 0.392& 0.078&  0.631&0.748 &3.6\\
				CBF~\cite{CBF} & 6.907 & 13.815 & 0.414& 0.319& 0.331& 0.563& 0.672&0.047\\
				WLS~\cite{WLS} & 6.821 & 13.642 & 0.490& 0.377& 0.223& 0.691& 0.930&0.85\\
				JSR~\cite{JSR_IF} &6.575  & 13.149 & 0.376& 0.223& 0.222& 0.601& 0.867&0.0033\\
				JSRSD~\cite{JSRSD} & 6.884 & 13.767 & 0.343& 0.199& 0.319& 0.539& 0.788&0.0027\\
				CSR~\cite{CSR} & 6.433 & 12.866 & \color{red}\bf 0.531 & 0.388& \color{red}\bf 0.021& \color{red}\bf 0.723& 0.906& 0.0074\\
				\hline
				\textbf{MFNet (Ours)} & \color{red}\bf 6.912& \color{red}\bf 13.825 & 0.425& 0.428 & 0.078 & 0.714 & 0.895 & \color{red}\bf 26.6 \\
				\hline
			\end{tabular}
		\end{center}
	\end{table*}
	
		\begin{table*}[t]
		\caption{Effectiveness of each component for JMMAC on the VOT19-RGBT, GTOT and RGBT234 datasets. The top three results are in {\color{red}\bf RED}, {\color{blue}blue} and {\color{green}green} fonts.}\label{tab-ablation}
		\footnotesize
		\begin{center}
			\begin{tabular}{c|ccc|cc|cc}
				\hline
				\multirow{2}{*}{Trackers} & \multicolumn{3}{c|}{VOT19-RGBT} & \multicolumn{2}{c|}{GTOT} & \multicolumn{2}{c}{RGBT234} \\
				
				& EAO~($\uparrow$) & A~($\uparrow$) & 	R~($\uparrow$)& MSR~($\uparrow$) & 	MPR~($\uparrow$) & MSR~($\uparrow$) & MPR~($\uparrow$)\\
				\hline
				JMMAC(B)-RGB & 0.3207 & 0.5909 & 0.6987 & 62.1 & 75.3 & 52.8 & 72.2\\
				JMMAC(B)-T & 0.3862 & 0.6452 & 0.7604 & 65.4 & 74.2 & 51.6 & 71.1\\
				\hline
				JMMAC(B)+MF-G & 0.4102~{\scriptsize(+2.50\%)} & \color{green}0.6502 & 0.7835 & 67.5~{\scriptsize(+2.1\%)}& 81.5~{\scriptsize(+6.2\%)}& 54.3~{\scriptsize(+1.5\%)} & 74.6~{\scriptsize(+2.4\%)}\\
				JMMAC(B)+MF-L & 0.4073~{\scriptsize(+2.11\%)} & 0.6387 & 0.7835 &67.7~{\scriptsize(+2.3\%)}&79.8~{\scriptsize(+4.5\%)}& 55.4~{\scriptsize(+2.6\%)} & 75.4~{\scriptsize(+3.2\%)}\\
				JMMAC(B)+MF & 0.4116~{\scriptsize(+2.54\%)} & 0.6465 & 0.7835 & 70.5~{\scriptsize(+5.1\%)} & 85.1~{\scriptsize(+9.8\%)}& 55.6~{\scriptsize(+2.8\%)} & 75.4~{\scriptsize(+3.2\%)}\\
				\hline
				JMMAC(B)+MF+CME & {\color{green}0.4198}~{\scriptsize(+0.82\%)} & \color{blue}0.6505 & \color{green}0.7953 & {\color{green}70.5}~{\scriptsize(+0.0\%)} & {\color{green}85.1}~{\scriptsize(+0.0\%)}& {\color{green}56.0}~{\scriptsize(+0.4\%)} & {\color{green}76.4}~{\scriptsize(+1.0\%)}\\
				\hline
				JMMAC(B)+MF+CME+TMP &  {\color{blue}0.4598}~{\scriptsize(+4.00\%)} & 0.6497 & \color{blue}0.8073& {\color{blue}70.9}~{\scriptsize(+0.4\%)} & {\color{blue}85.5}~{\scriptsize(+0.4\%)}& {\color{blue}56.4}~{\scriptsize(+0.4\%)} & {\color{blue}77.1}~{\scriptsize(+0.7\%)}\\
				\hline
				\textbf{JMMAC} & {\color{red}\bf0.4978}~{\scriptsize(+3.80\%)} & \color{red}\bf0.6597 & \color{red}\bf0.8235 & {\color{red}\bf73.2}~{\scriptsize(+2.3\%)} & {\color{red}\bf90.1}~{\scriptsize(+4.6\%)}& {\color{red}\bf 57.3}~{\scriptsize(+0.9\%)} & {\color{red}\bf 79.0}~{\scriptsize(+1.9\%)}\\
				\hline
			\end{tabular}
		\end{center}
	\end{table*}

\noindent{\textbf{Effectiveness of Different Components.}} To provide a thorough analysis of each component, we compare several variants of our JMMAC tracker, including: 
(1) JMMAC(B)-RGB: the baseline method with only RGB modality; 
(2) JMMAC(B)-T: the baseline method with only thermal modality; 
(3) JMMAC(B)+MF-G: JMMAC(B) only with Global MFNet; 
(4) JMMAC(B)+MF-L: JMMAC(B) only with Local MFNet; 
(5) JMMAC(B)+MF: JMMAC(B) with MFNet (i.e., the combination of both global and local MFNets); 
(6) JMMAC(B)+MF+CME:  JMMAC(B) with multimodal fusion (MF) and camera motion estimation (CME) modules; 
(7) JMMAC(B)+MF+CME+TMP:  JMMAC(B) with multimodal fusion (MF), camera motion estimation (CME) and 
target motion prediction (TMP) modules; 
(8) JMMAC: our final model, i.e., JMMAC(B)+MF+CME+TMP+BR. `BR' means that the tracker exploits the detection scheme to refine the output bounding boxes if necessary.
The detailed comparison results on all three datasets are reported in Table~\ref{tab-ablation}. 
Each component substantially contributes to the final JMMAC tracker and 
gradually improves the tracking performance. Both global and local fusion networks can effectively improve the baseline. 
Compared with the data shown in Table~\ref{tab-vot19-rgbt}, our tracker performs better than other competing algorithms when used with multimodal fusion only (see JMMAC(B)+MF \emph{vs} SiamDW\_T), which shows the effectiveness of the proposed MFNet. 
Since the sequences are captured by still camera in GTOT, CME module does not influence the tracking result.

\noindent{\textbf{MFNet analysis on image fusion.}} We find that the goals of image fusion and multimodal fusion in tracking are similar. In tracking, we assume that the more possible the target is, the more salient information the modality contains, while the image fusion aims to combine the visible and thermal images while preserving both thermal radiation and detailed texture information. To this end, our learned MFNet can be also generalized to deal with the image fusion task (even without fine-tuning on related datasets).
As discribed in Section~\ref{MFNet}, we send the RGB-T images to MFNet and obtain the fused image via ${\bf I}_F = {\bf W}_F \times {\bf I}_{RGB} + (1-{\bf W}_F) \times {\bf I}_{T}$.
We compare our MFNet with other algorithms using the dataset presented in~\cite{FusionGAN}, which includes $41$ pairs of testing images collected from TNO and INO datasets. 
The TNO dataset\footnote{\href{https://figshare.com/articles/TNO_Image_Fusion_Dataset/1008029}{https://figshare.com/articles/TNO\_Image\_Fusion\_Dataset/1008029}} contains image sequences registered with different multiband camera systems, recording multi-spectral nighttime imagery of different military relevant scenarios. The INO dataset\footnote{\href{https://www.ino.ca/en/video-analytics-dataset/}{ https://www.ino.ca/en/video-analytics-dataset/}}, constructed by the National Optics Institute of Canada, consists of several image pairs and sequences captured in different conditions.

In this paper, we choose eight common metrics for evaluation, including entropy (EN)~\cite{EN}, mutual information (MI)~\cite{MI}, edge information (Qabf)~\cite{Qabf}, feature mutual information (FMI)~\cite{FMI}, fusion artifacts (Nabf)~\cite{Nabf}, structural similarity index measure (SSIM)~\cite{SSIM}, multi-scale SSIM (MS\_SSIM)~\cite{MSSSIM}, and inference speed (frame per second, FPS).
The comparison results in Table~\ref{table_IF} indicate that our MFNet achieves very competitive performance with real-time performance. 
Our MFNet performs consistently better than FusionGAN (the second fast method) with speeds seven time faster. 
Furthermore, we show 8 pairs of fusion results to validate that our MFNet can also achieve very competitive results on the image fusion task. The testing images are from previous work~\cite{FusionGAN}. As shown in Figure~\ref{fig:qualitiveimagefusion}, MFNet can fuse complementary information from both modalities while retaining more detail information in the visible modality.

\begin{figure}
	\centering
	\includegraphics[width=1.0\linewidth]{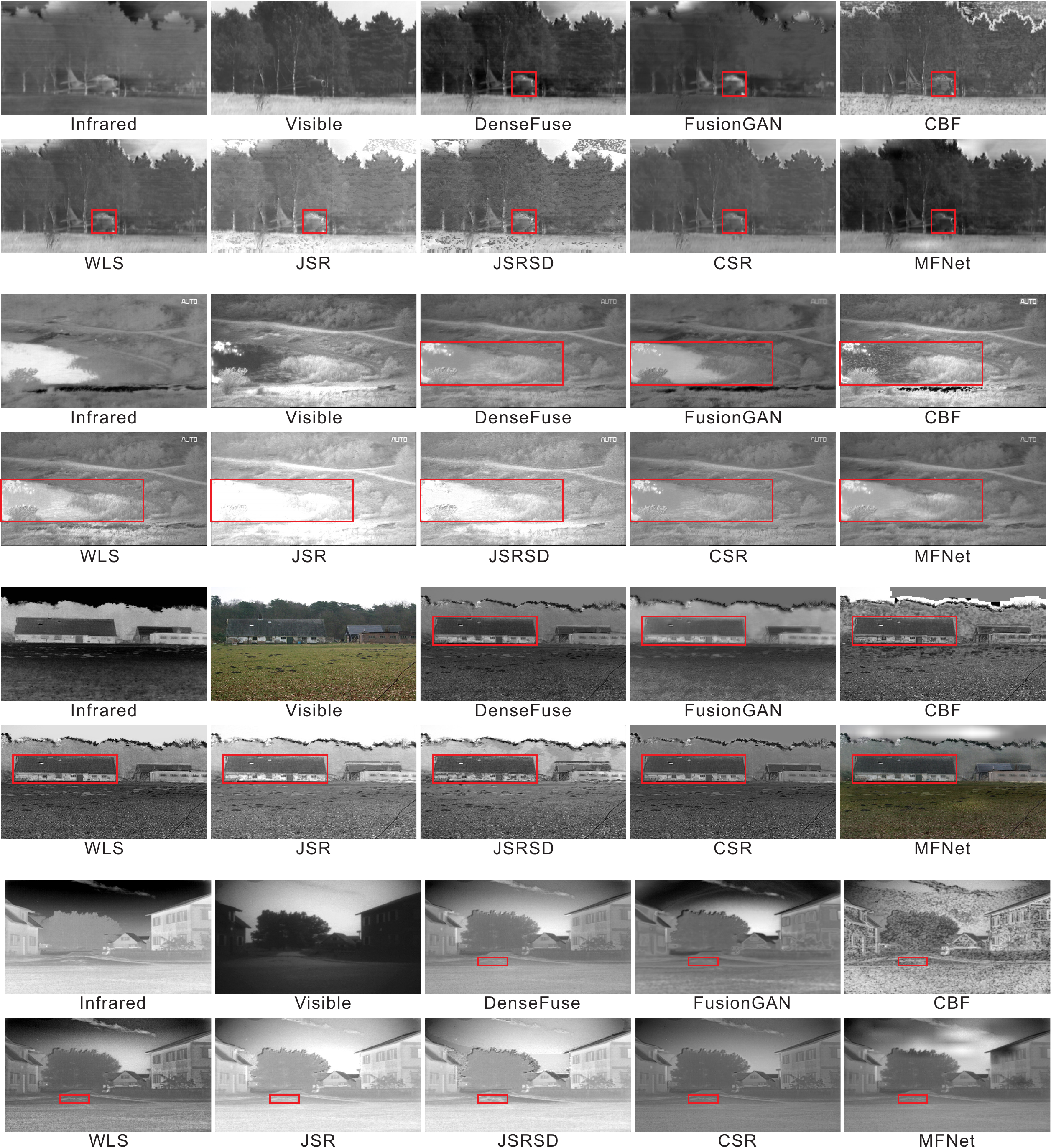}
	\caption{The visualization results on image fusion task.}
	\label{fig:qualitiveimagefusion}
\end{figure}

\begin{table}[t]
	\caption{Comparisons of different camera motion models. Here, we adopt the JMMAC(B)+MF variant as the baseline.}   
	\label{tab-cme}
	\footnotesize
	\begin{center}
		\begin{tabular}{cccc}
			\hline
			Motion models & EAO~($\uparrow$) & A~($\uparrow$) & R~($\uparrow$)\\
			\hline
			JMMAC(B)+MF & 0.4116 & 0.6465 & 0.7835\\
			JMMAC(B)+MF+Translation & 0.3825 & {\color{red}\bf0.6527} & 0.7529\\
			JMMAC(B)+MF+Similarity & {\color{green}0.4163} & 0.6476 & {\color{red}\bf0.8073}\\
			JMMAC(B)+MF+Affine  &{\color{red}\bf0.4198} & {\color{green}0.6505} & {\color{blue}0.7953}\\
			JMMAC(B)+MF+Projective & {\color{blue}0.4136}& {\color{blue}0.6516}& {\color{green}0.7835}\\
			\hline
		\end{tabular}
	\end{center}
\end{table}
\noindent{\textbf{Analysis of fusion methods.}} We also implement some variants with earlier or late fusion manners, and find that the proposed MFNet performs the best. We apply 5 other fusion methods to validate the strength of MFNet, which contains both early fusion and late fusion methods. (1) Merge: layer-wise add the features in RGB and thermal modalities. (2) Concatenate: layer-wise concatenate the features in RGB and thermal modalities. (3) Concatenate+PCA: We apply feature concatenation followed by Principal Component Analysis~(PCA) operation. (4) Intensity-based fusion: we assume that the temperature of object is constant and we calculate the intensity of target $i_1$ in the first frame. we add a penalty to the response expressed as,
\begin{figure}[t]
	\centering
	\includegraphics[width=1.0\linewidth]{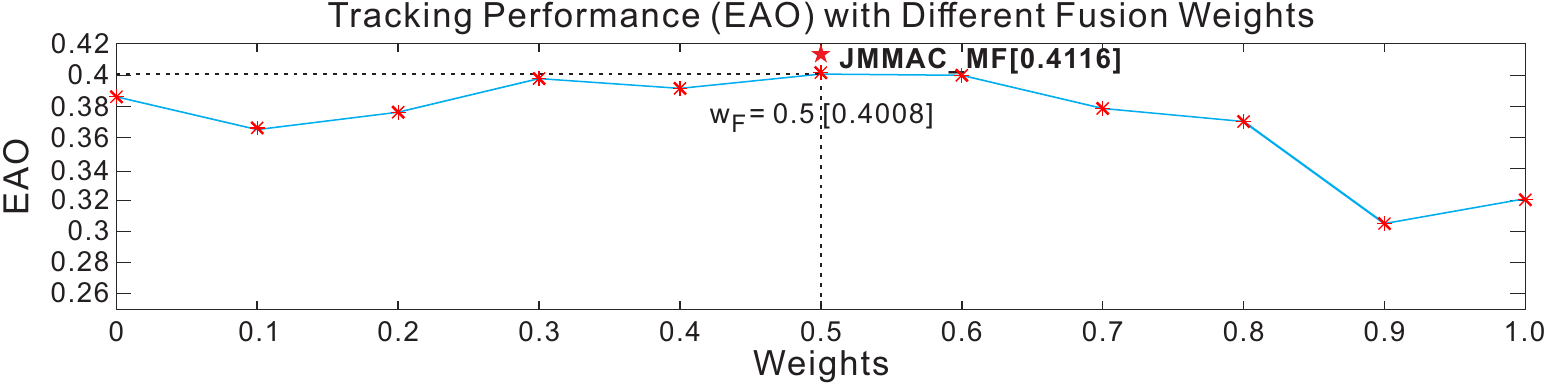}
	\caption{The tracking performance with different fusion weights $w$ in VOT19-RGBT.}
	\label{fig:mfsearchparametersupplementary}
\end{figure}
\begin{equation}
{\bf R}_{F} =  \frac{1}{2}{\bf P} \times ({\bf R}_{RGB} + {\bf R}_{T})
\end{equation}
\begin{equation}
s.t.~~ {\bf P}(i,j) = min(\frac{i_t(i,j)}{i_1},\frac{i_1}{i_t(i,j)})
\end{equation}
\begin{table*}[t]
	\caption{Fusion methods analysis on VOT19-RGBT. Both early fusion and late fusion methods are included for evaluation. Compared with different type of fusion methods, our MFNet obtains the best performance.}\label{MF}
	\footnotesize
	\begin{center}
		\begin{tabular}{cccccc}
			\hline
			Fusion type & Fusion method & Available modality & EAO~($\uparrow$) & A~($\uparrow$) & 	R~($\uparrow$)\\
			\hline
			\multirow{2}{*}{Single modality} & - & RGB & 0.3207& 0.5909&0.6987\\
			&-&Thermal & 0.3862 & \color{green}0.6452 & 0.7604 \\
			\hline
			\multirow{3}{*}{Early fusion} & Merge & RGB \& Thermal &0.3734 &0.6350 &0.7567\\
			& Concatenate & RGB \& Thermal &\color{green}0.3976 &0.6387 &0.7567 \\
			& Concatenate + PCA & RGB \& Thermal &\color{blue}0.3980 &0.6373 &\color{green}0.7681 \\
			\hline
			\multirow{3}{*}{Late fusion} & Intensity-based fusion& RGB \& Thermal & 0.3870&0.6324 &\color{blue}0.7796\\
			& Tracking-quality-based fusion & RGB \& Thermal & 0.3647&\color{red}0.6480 &0.7454\\
			& \bf {MFNet~(Ours)} & RGB \& Thermal & \color{red}0.4116 &\color{blue}0.6465 & \color{red}0.7835 \\
			\hline
		\end{tabular}
	\end{center}

\end{table*}
where $(i,j)$ denotes the location in response map and $i_t$ denotes the intensity of target in the $t$-th frame. (5) Tracking-quality-based fusion: we fuse the responses with the guidance of tracking quality described in \cite{FCLT} and the final response can be expressed as, 
\begin{equation}
{\bf R}_{F} = \frac{q_{RGB}}{q_{T}+q_{RGB}}\times{\bf R}_{RGB} + \frac{q_{T}}{q_{T}+q_{ RGB}}\times{\bf R}_{ T}
\end{equation}
where $q_{RGB}$ and $q_{T}$ are tracking quality of RGB and thermal modalities calculated by Equation~(5) in our submission, (6) MFNet: our proposed approach. Also, we add the comparison between the trackers solely with RGB and thermal information. From Table~\ref{MF}, all fused methods can improve tracking performance except for merge and tracking-quality-based fusion. This may be caused by the discrepancy of data from different modalities and the uncertainty of tracking quality. MFNet outperforms all other fusion methods in a large margin. Late fusion, which can extract features in different layers and fuse them in various scales, is more flexible than early fusion, that fusion operation only can be applied to feature in the same size. Furthermore,  We fuse the responses via a weighted linear combination, i.e., ${\bf R}_F = w \times {\bf R}_{RGB} +(1-w) \times {\bf R}_T$. In Figure~\ref{fig:mfsearchparametersupplementary}, we enumerate fusion weight $w$ from 0 to 1 with an interval 0.1 and report the tracking performance~(EAO) on VOT19-RGBT.

\noindent{\textbf{Parameter Robustness Analysis of JMMAC}}. Since both online information~(quality of response) and offline information~(template in the initial frame) are considered to jointly determine which cue is used for tracking, we argue that our target motion prediction~(TMP) module is very robust against parameter perturbation and the parameters are not over-fitting to specific dataset. To fully validate this, we enumerate the all the parameters with interval 2~(for $t_{diff}$, we set the interval to 1)and report their EAOs in VOT19-RGBT. As shown in Figure \ref{fig:parameter_robustness}, compared with the tracker without TMP module~(JMMAC-TMP for short), JMMAC with TMP achieves performance promotion with a large parameter range.

\begin{figure}
	\centering
	\includegraphics[width=1.0\linewidth]{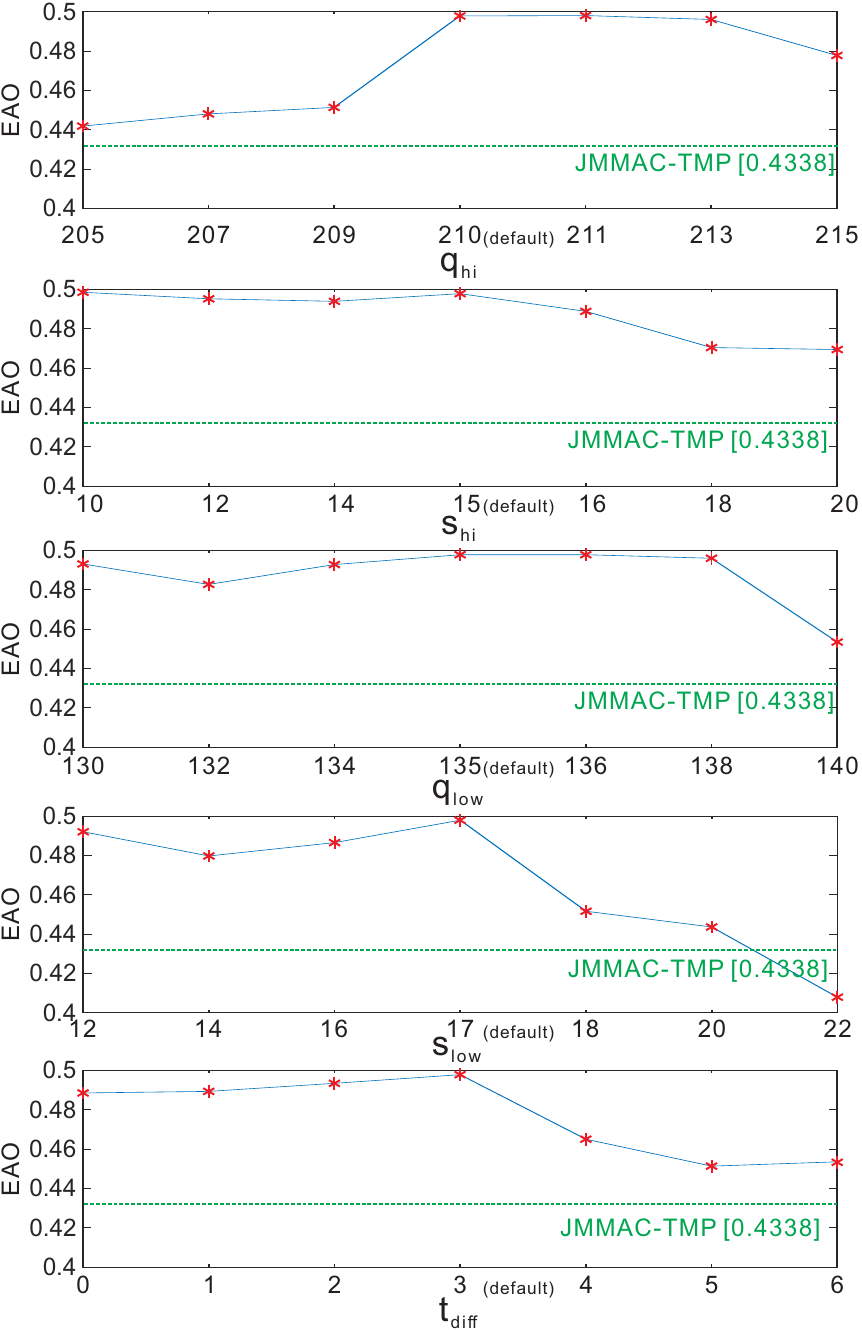}
	\vspace{-8mm}
	\caption{Parameter robustness analysis. We enumerate the parameters with the interval 2~(especially, for $t_{diff}$, we set the interval to 1) individually. Our tracker switcher is robust against parameter perturbation and works well in a large range.}
	\label{fig:parameter_robustness}
\end{figure}

\noindent{\textbf{Different Camera Motion Models.}} We also test the effects of different camera motion 
models, including translation, similarity, affine, and projective transformations between frames. 
The comparison results are summarized in Table~\ref{tab-cme}. All camera motion models, except for 
translation transformation, can improve the tracking performance. 
Among those models, the affine transformation performs best and serves as our final model in camera motion estimation module. 

\begin{table}[t]~\label{speed}
	\caption{Speed analysis for each component in JMMAC.}   
	\scriptsize
	\vspace{-5mm}
	\begin{center}
		\begin{tabular}{ccccccc}
			\hline
			Module & JMMAC(B) & MFNet & CME & TMP & BR & JMMAC \\
			\hline
			Time (sec.) & 0.1160 & 0.0124 & 0.0551 & 0.0737 & 0.0415 & 0.2664\\
			\hline
		\end{tabular}
	\end{center}
	\vspace{-5mm}
\end{table}

\noindent{\textbf{Speed Analysis.}} We conduct speed analysis for JMMAC and show the average time cost for each component in Table.~\ref{speed}. Our tracker approximately runs at 4 FPS and the main time cost is from the appearance tracker~(JMMAC(B)) with the deep feature. Our proposed MFNet is efficient to fuse the multimodal information and provide a final response and the target motion prediction(TMP) and camera motion estimation(CME) machanisms do not bring significant speed decline. The box regression with a real-time YOLOv2 is applied to refine the bounding box, whose computation can be negligible.

\section{Conclusion}
In this study, we propose a novel JMMAC method for robust RGB-T tracking. 
Our method effectively exploits both appearance and motion cues in dealing with the RGB-T tracking task. 
For the appearance information, we develop a novel MFNet to infer the fusion weight maps of both RGB and 
thermal modalities, resulting in a much reliable response map and a robust tracking performance. 
The experiments demonstrate that our MFNet is not only suitable for improving the tracking accuracy but also 
competitive in handling the image fusion task.
For the motion information, we attempt to jointly consider camera motion and target motion, enabling the 
tracker to become much more robust when the appearance cue is unreliable. 
Extensive results on GTOT, RGBT234, and VOT19-RGBT datasets show that the proposed JMMAC tracker 
achieves remarkably better performance than other state-of-the-art algorithms.

\bibliographystyle{abbrv}
\bibliography{JMMAC}

\end{document}